# Detecting Vehicle Type and License Plate Number of different Vehicles on Images

Aashna Ahuja
Mukesh Patel School of Technology
Management and Engineering
Mumbai India
aashna.ahuja1721@nmims.edu.in

Arindam Chaudhuri
Mukesh Patel School of Technology
Management and Engineering
Mumbai India
arindam.chaudhuri@nmims.edu

**Abstract:** With ever increasing number of vehicles, vehicular tracking is one of the major challenges faced by urban areas. In this paper we try to develop a model that can locate a particular vehicle that the user is looking for depending on two factors 1. the Type of vehicle and the 2. License plate number of the car. The proposed system uses a unique mixture consisting of Mask R-CNN model for vehicle type detection, WpodNet and pytesseract for License Plate detection and Prediction of letters in it.

**Keywords:** Vehicle Detection, Computer Vision, Image Classification, Vehicle Tracking, R-CNN

## 1 Introduction

Detecting and classifying objects is a rapidly expanding field. It can track cars, creatures, dents, defects, locations, and other structures in the physical world, among other things. This paper explains how different models, such as Mask R-CNN, WpodNet, and Tesseract, can be integrated and used in real-world scenarios for various applications. The individual models used have been proven to be best in their work therefore, combining them together to build a model helps us to get a good accuracy and faster detection. The final model can then be used for scenarios like tracking vehicle in parking lots, on highways. We can also search for different vehicles based on certain criteria like vehicle type and its license plate. Various libraries like TensorFlow, keras, cv2, matplotlib and os were used to build these models.

The aim of this project is to merge various models in order to address a larger real-world challenge. The aim was to construct a model that could be used to find a vehicle by combining the two variables of type and licence plate.

This paper is organised as follows. In section 2 related work is presented. This is followed by datasets used and related models in section 3. In section 4 methodology is highlighted. Section 5 presents the results. In section 6 conclusion is given.

## 2 Related Work

Various research work has been published in object detection and classification domain. Based on these three models viz Mask R-CNN, WpodNet and Tesseract were chosen to be used for making the final model. In License Plate Detection and Recognition in Unconstrained Scenarios [1] it describes the use of WpodNet for recognition and detection of license plates in different scenarios. It helps us to gain insight into the WpodNet architecture and working.

Real Time License Plate Recognition from Video Streams using Deep Learning [2] gives us a base to start or license plate detection work. It helps us understand the process of recognising a plate and tries to reach the accuracy of 91%. Fast and Improved Real-Time Vehicle Anti-Tracking System [3] is another useful paper that applies the vehicle detection models to a real-world scenario. It uses harr cascade models to locate the vehicle.  All the prior work that has been done in this field either detects the vehicle majorly cars in the image/video and does a classification [4]. For the license plate detection in various papers single model has been used.   Therefore, this work is an extension to all previous papers.

## 3 Datasets Used and Related Models

For Mask R-CNN model, 2560 images were manually annotated and used for training, while 100 images were used for testing. The WpodNet model was trained on more than 1000 different images for each of the 36 categories [0-9, A-Z]. All the photos collected were of jpg format and were taken from open-source websites.

### 3.1 Mask R-CNN

Mask R-CNN is state-of-the-art build using deep neural network and performs best for solving Instance segmentation problems. Instance segmentation means breaking the image into various parts so that it is easier to recognise and identify the object contained in the image.

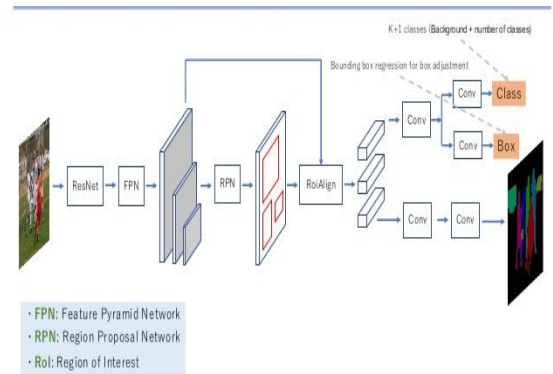

Fig. 1:  Mask R-CNN Architecture

Mask R-CNN is an extension of Faster RCNN and works on ResNet backbone. The Faster R-CNN operates in two stages: in the first stage it tries to locate the object with the help of region proposal network (RPN). Region proposals are basically regions in the feature map which contain the object that we are looking for. In the second stage the model predicts the bounding boxes and class objects. Faster R-CNN predicts object class and bounding boxes. Whereas Mask R-CNN has an additional branch for predicting segmentation masks on individual Region of Interest (ROI). Mask RCNN in the end returns a class, a bounding box and a mask as shown in Fig. 1.

### 3.2 WpodNet

WpodNet stands for Warped Planar Object Detection Network. After training this network it can learn to detect different LPs in different scenarios.  The WpodNet was developed by using information related to YOLO, SSD and STN. Yolo and SSD are known for detecting multiple objects very fast and STN is known for handling non regular objects. For detecting the LP first an area of high interest is located where the LP would most likely be. Then in the area of interest the object is detected. Fig. 2 shows whole process.

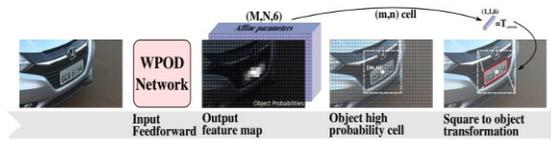

Fig. 2: WpodNet Detection Method

### 3.3 Pytesseract

Python-tesseract is an optical character recognition tool for python. That is, it recognizes and reads the text embedded in images. Python-tesseract is a small, wrapped version of Google's Tesseract-OCR Engine [5]. It supports all file formats provided by Pillow and other imaging libraries, including jpeg, png, gif, bmp, tiff, and others.

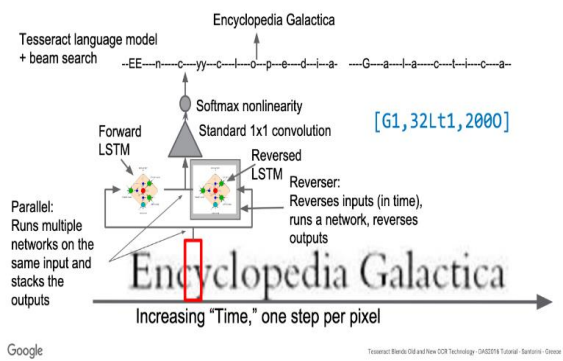

Fig. 3: Pytesseract in action

Fig. 3 shows structure of pytesseract model. It is made up of various LSTM Structures and has three main stages for locating and predicting the letters on a license plate: word finding, line finding, and character classification. Lines are arranged into blobs in the Word finding step, and then the lines are scanned for proportion text. Lines are then broken down into characters based on the spaces.

A two-step procedure is then used to recognize the characters: The tesseract first attempts to identify each term, and then passes the data to an adaptive classifier as training data. The adaptive classifier is then given the opportunity to identify text farther down the page with greater accuracy.

## 4 Methodology

Here we will examine the steps used for building the new model. The first stage consisted of annotating the images and training the Mask R-CNN model on our custom dataset. In total 2650 images were annotated using an open-source tool (VGG Annotator). Rectangular shape was used instead of polygons. Fig. 4 below shows the example of how annotations were done.

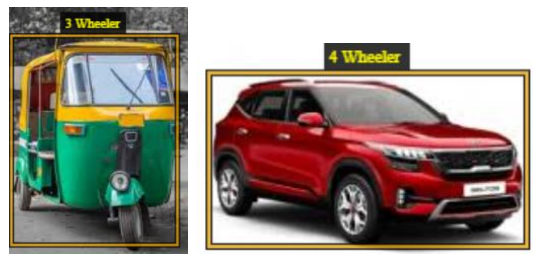

Fig. 4: Annotated images

Next the Mask R-CNN model was trained for four different categories that are [6], [7]:

(a) 2-Wheeler (Bikes, Scooty and Cycles)
(b) 3-Wheeler (Autos and Vans)
(c) 4-Wheeler (Cars, Vans and Jeeps)
(d) >4 Wheeler (Trailor, Trucks and Buses)

The model gave us an F1 score of 0.72399 on training and 0.68 on testing data. Fig. 5 shows the output of trained model.

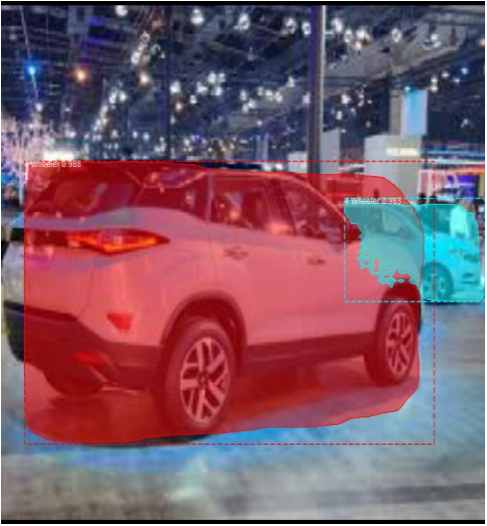

Fig. 5: Vehicle Type Detection using Mask R-CNN

The second stage was divided in two parts. In the first part the task was to get the license plate of a car and correctly classify the letter of the license plate. The below Fig. 6 and Fig. 7 shows output of WpodNet model that has an accuracy of >0.90.

The second part of this stage was to use pytesseract to predict the letters of the license plate accurately. The license plate image extracted from WpodNet was fed as input to the pytesseract model. This made the process faster and more accurate.

The third and most important step was to take user input for type, license plate and the image. The model then matches the user input and model outputs and tells us whether the vehicle that we are looking for has been found or not.

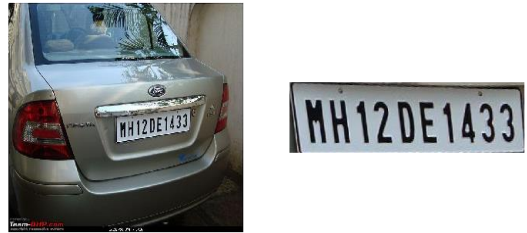

Fig. 6: Number Plate Extraction using WpodNet

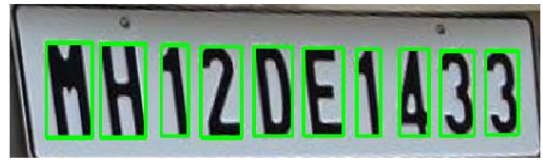

Fig. 7: Letter Detection using WpodNet

## 5 Results

The final model was a hybrid of the three models described above. For car identification, we used Mask R-CNN and for license plate detection, we use WpodNet and pytesseract. In the final model, the Mask R-CNN gave an F1 score of 0.72399 on training dataset, and for testing, it gave an F1 score of 0.68. To look for vehicle, user types the vehicle type and license plate number that they are searching for, and the model returns whether it was effective in locating the vehicle or not. This work can be applied to real-world situations such as locating a stolen vehicle or robbers in a car using cameras using a website/application, or recording vehicle entrance and departure in malls, institutes, and other parking lots to determine the number of open/available parking spaces.

## 6 Conclusion

In this work, a combination of all the three models has been used to build a high-performance system. The system is able to find the vehicle as per the user's input. This model in future can be further improved by improving the F1 score of Mask R-CNN model and for the license plate detection the confusion between letter 'M' and 'N' could be cleared. And finally, this can also be tried on live video captured from cameras on highway.